\documentclass[11pt,letterpaper,logo,twocolumn]{style}

\usepackage[numbers]{natbib}
\usepackage{graphicx}
\usepackage{booktabs}
\usepackage{amsmath,amsfonts,amssymb}
\usepackage{subcaption}
\usepackage{multirow}
\usepackage{colortbl}
\usepackage{listings}
\usepackage{xparse}
\usepackage{fontawesome5}
\usepackage{threeparttable}

\graphicspath{{./}{fig/}{figures/}{plot/}{pdf/}{table/}}
\usepackage{amsthm}
\tcbuselibrary{skins,breakable}
\tcbuselibrary{listingsutf8}
\usepackage{titletoc}
\usepackage{pifont}
\usepackage{mathtools}
\usepackage{bbm}
\usepackage{makecell}
\usepackage{adjustbox}
\usepackage{hyperref}
\usepackage[normalem]{ulem} 
\usepackage{amsmath}
\usepackage{comment}
\usepackage{booktabs}
\usepackage{tabularx}
\usepackage{ragged2e}
\usepackage{pifont}

\usepackage{times}
\usepackage{latexsym}
\usepackage[edges]{forest}
\usepackage[T1]{fontenc}
\usepackage[utf8]{inputenc}
\usepackage{microtype}
\usepackage{inconsolata}
\usepackage{tcolorbox}
\usepackage{subcaption}
\usepackage{lipsum}
\usepackage{multirow}
\usepackage{array}
\newcolumntype{C}[1]{>{\raggedright\arraybackslash}p{#1}}
\definecolor{softpurple}{RGB}{229,224,236}  
\definecolor{softrose}{RGB}{244,224,232} 
\definecolor{softpink}{RGB}{248,226,234} 
\definecolor{softred}{RGB}{242,220,219}     
\definecolor{softcoral}{RGB}{250,224,222} 
\definecolor{softpeach}{RGB}{255,230,220} 
\definecolor{softorange}{RGB}{255,236,214} 
\definecolor{softgold}{RGB}{249,239,206} 
\definecolor{softyellow}{RGB}{255,242,204}  
\definecolor{softbeige}{RGB}{240,232,220} 
\definecolor{softsand}{RGB}{244,238,230} 
\definecolor{softolive}{RGB}{235,241,222} 
\definecolor{softsage}{RGB}{229,239,224} 
\definecolor{softgreen}{RGB}{226,239,218}   
\definecolor{softmint}{RGB}{221,242,234} 
\definecolor{softteal}{RGB}{208,232,230} 
\definecolor{softcyan}{RGB}{220,245,245} 
\definecolor{softperiwinkle}{RGB}{224,230,246}
\definecolor{softsky}{RGB}{217,235,247} 
\definecolor{softblue}{RGB}{220,230,242}    
\definecolor{softbluegray}{RGB}{226,232,238} 
\definecolor{softgray}{RGB}{240,240,240}     
\definecolor{magenta}{RGB}{227, 0, 140}
\definecolor{orange}{RGB}{234, 67, 0}

\newcommand{\revise}[1]{}

\newtcolorbox{takeaway}[1]{
    colback=softbluegray,    
    colframe=softbluegray,  
    fonttitle=\bfseries,      
    coltitle=black,           
    attach title to upper,    
    after title={\ },         
    sharp corners,            
    boxrule=0.pt,             
    left=5pt, right=5pt, top=5pt, bottom=5pt, 
    title={#1},               
    fontupper=\small
}

\usepackage[utf8]{inputenc}
\usepackage[T1]{fontenc}
\usepackage{geometry}

\usepackage{mathptmx} 
\usepackage{amsmath, amssymb}
\usepackage{booktabs} 

\usepackage[dvipsnames]{xcolor}

\newtcolorbox{manifestation}{
    enhanced,
    breakable,
    frame hidden,                     
    borderline west={2.5pt}{0pt}{gray!70}, 
    colback=gray!5,                   
    coltitle=gray!90,                 
    fonttitle=\bfseries\sffamily\small,
    title={Empirical Manifestation},  
    attach title to upper={\par\vspace{3pt}}, 
    fontupper=\small\itshape,         
    parbox=false,
    before skip=10pt,
    after skip=10pt
}

\newtcolorbox{summarybox}[1][]{
    enhanced,
    colback=MidnightBlue!5,      
    colframe=MidnightBlue!75,    
    coltitle=white,              
    fonttitle=\bfseries\sffamily,
    title={#1},
    arc=2pt,                     
    boxrule=0.8pt,               
    drop fuzzy shadow=gray!30,   
    left=8pt, right=8pt, top=8pt, bottom=8pt,
    before skip=15pt
}

\title{Reasoning emerges from constrained inference manifolds in large language models}
\runningtitle{Reasoning emerges from constrained inference manifolds in large language models}
\PublicDate{2026-5-1}
\newcommand{\projectlead}{\textsuperscript{\ddag}}

\author{%
  {\Authfont \small
    \textbf{Yanbiao Ma}\textsuperscript{1} \quad
    \textbf{Fei Luo}\textsuperscript{1} \quad
    \textbf{Linfeng Zhang}\textsuperscript{2,3} \quad
    \textbf{Chuangxin Zhao}\textsuperscript{2} \quad 
    \textbf{Mingxuan Wang}\textsuperscript{1} \quad
    \textbf{Yinan Wu}\textsuperscript{2} \quad
    \textbf{Zhe Qian}\textsuperscript{1} \quad
    \textbf{Yang Lu}\textsuperscript{4} \quad
    \textbf{Long Chen}\textsuperscript{3} \quad
    \textbf{Zhao Cao}\textsuperscript{1} \quad
    \textbf{Xiaoshuai Hao}\advisor\projectlead \textsuperscript{3} \quad
    \textbf{Ji-Rong Wen}\advisor \textsuperscript{1} \quad
    \textbf{Jungong Han}\advisor\textsuperscript{2} \quad
  }\\
  {\Affilfont
    \textsuperscript{1} Renmin University of China \quad
    \textsuperscript{2} Tsinghua University \quad
    \textsuperscript{3} Xiaomi EV \quad
    \textsuperscript{4} Xiamen University \quad
    \\
    \advisor\ Corresponding authors \quad \projectlead\ Project leader \quad
    \faEnvelope\: \texttt{ybma1998@ruc.edu.cn,
haoxiaoshuai@xiaomi.com,
jrwen@ruc.edu.cn,
jghan@tsinghua.edu.cn}
  }
}

\begin{document}

\begin{abstract}

Reasoning in large language models is predominantly evaluated through labeled benchmarks, conflating task performance with the quality of internal inference.
Here we study reasoning as an intrinsic dynamical process by examining the evolution of internal representations during inference.
We find that inference-time dynamics consistently self-organize into low-dimensional manifolds embedded within high-dimensional representation spaces.
we find that such geometric compression, although pervasive, is not sufficient for stable or reliable reasoning.
Instead, effective reasoning dynamics emerge within a constrained structural regime characterized by three conditions: adequate representational expressivity, spontaneous manifold compression, and preservation of non-degenerate information volume within the compressed subspace.
Models outside this regime exhibit characteristic pathological inference dynamics.
Based on these insights, we introduce a unified, label-free diagnostic computed solely from internal dynamics.
These findings suggest that reasoning in LLMs is fundamentally governed by geometric and informational constraints, offering a complementary framework to benchmark-centric assessment.
\end{abstract}

\maketitle


\begin{figure*}[!ht]
\centering
\includegraphics[width=0.98\linewidth]{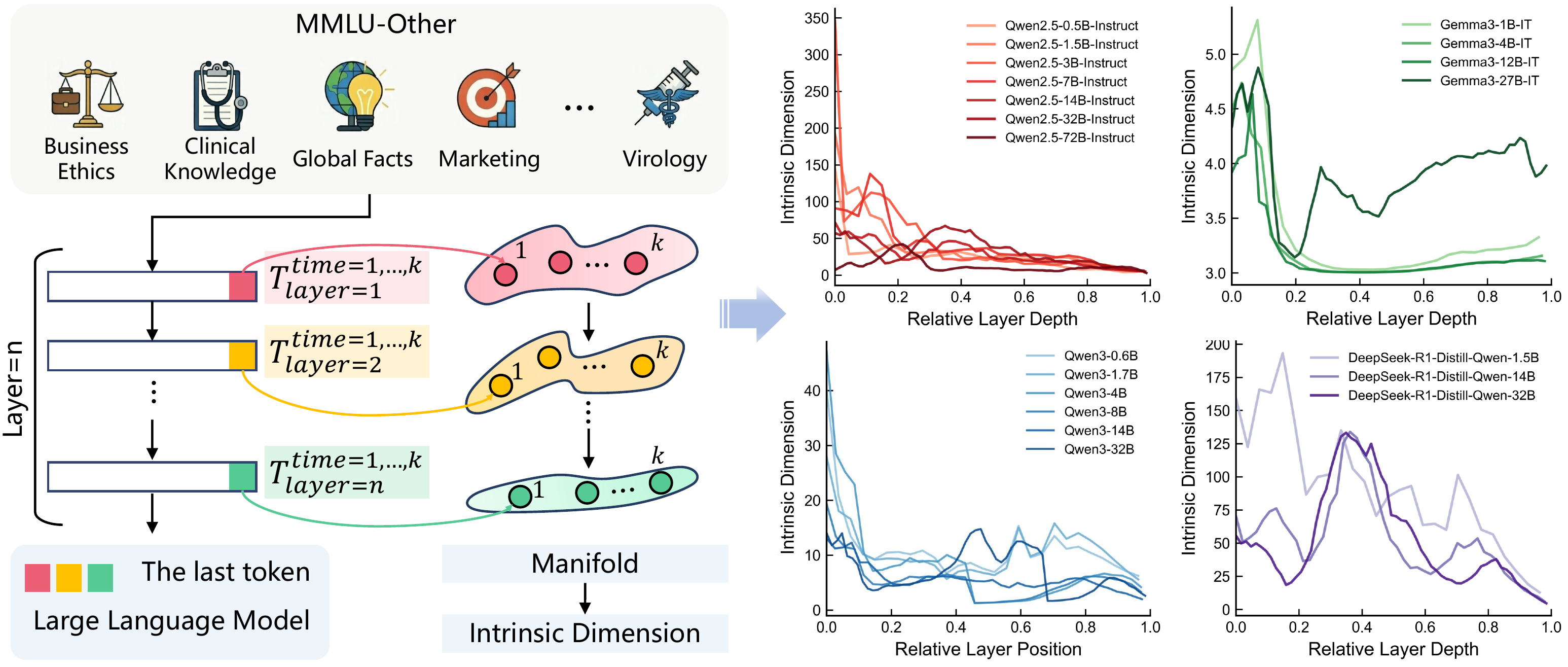}
\caption{\textbf{Inference-time reasoning dynamics self-organize into low-dimensional manifolds.} Layer-wise evolution of intrinsic dimensionality (ID) for inference-time representation trajectories across representative large language models under generic cognitive stimulation. Despite the high ambient dimensionality of the embedding space, ID rapidly decreases with network depth and stabilizes at values orders of magnitude smaller than the embedding dimension, indicating that inference dynamics become confined to a compact manifold.}
\label{fig1}
\end{figure*}


Large language models have shown remarkable improvements in reasoning across mathematics \cite{wei2022chain,kojima2022large}, science, and commonsense domains \cite{tang2026multimodal,hurst2024gpt,xiao2025densing}.
Yet reasoning ability is still predominantly evaluated using labeled benchmarks and task accuracy \cite{srivastava2023beyond,liang2022holistic}, implicitly treating reasoning as an opaque input–output mapping \cite{bubeck2023paper,mondorf2024beyond}.
Such evaluations conflate internal inference quality with dataset alignment and prompting strategies \cite{cao2025toward}, offering limited insight into how reasoning is internally realized or why models with similar accuracy can differ substantially in robustness and generalization.

To address this limitation, we study reasoning from an internal perspective \cite{elhage2022toy,geva2023dissecting,ciceri2024inversion}, characterizing it as a dynamical process unfolding in representation space during inference \cite{dreyer2025mechanistic}.
Rather than focusing on correctness, we analyze how internal representations evolve when models are engaged by generic cognitive stimuli, independent of task-specific supervision.

Across model families \cite{bai2025qwen25vltechnicalreport,yang2025qwen3,gemmateam2025gemma3technicalreport,guo2025deepseek}, scales, and prompts \cite{wang2024mmlu}, \textbf{we observe a striking and consistent regularity:} inference-time representations spontaneously collapse onto extremely low-dimensional trajectories, despite residing in highly expressive embedding spaces.
These trajectories indicate that reasoning dynamics are effectively confined to low-dimensional manifolds embedded within the ambient representation space.
This dimensional collapse is stimulus-induced, reproducible across runs, and emerges during inference rather than being imposed by architectural constraints.
However, low dimensionality alone does not guarantee robust reasoning.
Models with similarly compressed trajectories can exhibit sharply different inference behavior, and overly aggressive compression can lead to brittle or information-poor dynamics.

These observations indicate that robust reasoning occupies a narrow admissible regime of internal structure.
Healthy reasoning requires sufficient representational expressivity, spontaneous low-dimensional organization of inference dynamics, and non-degenerate information flow within the compressed manifold.
Violating any of these constraints leads to characteristic pathological regimes, including diffuse exploration, representational crowding, or degenerate collapse.

In this work, we uncover intrinsic constraints and introduce a unified, label-free diagnostic computed solely from inference-time representations.
Without relying on task supervision, this measure exposes whether internal trajectories reside within a geometrically and informationally admissible regime for stable reasoning. Applied across contemporary large language models, it consistently separates structurally coherent inference from pathological dynamics and predicts downstream robustness despite remaining entirely task-agnostic.
Together, our results reframe reasoning evaluation in terms of internal geometry and information flow rather than task accuracy alone.

\begin{figure*}[!ht]
\centering
\includegraphics[width=0.98\linewidth]{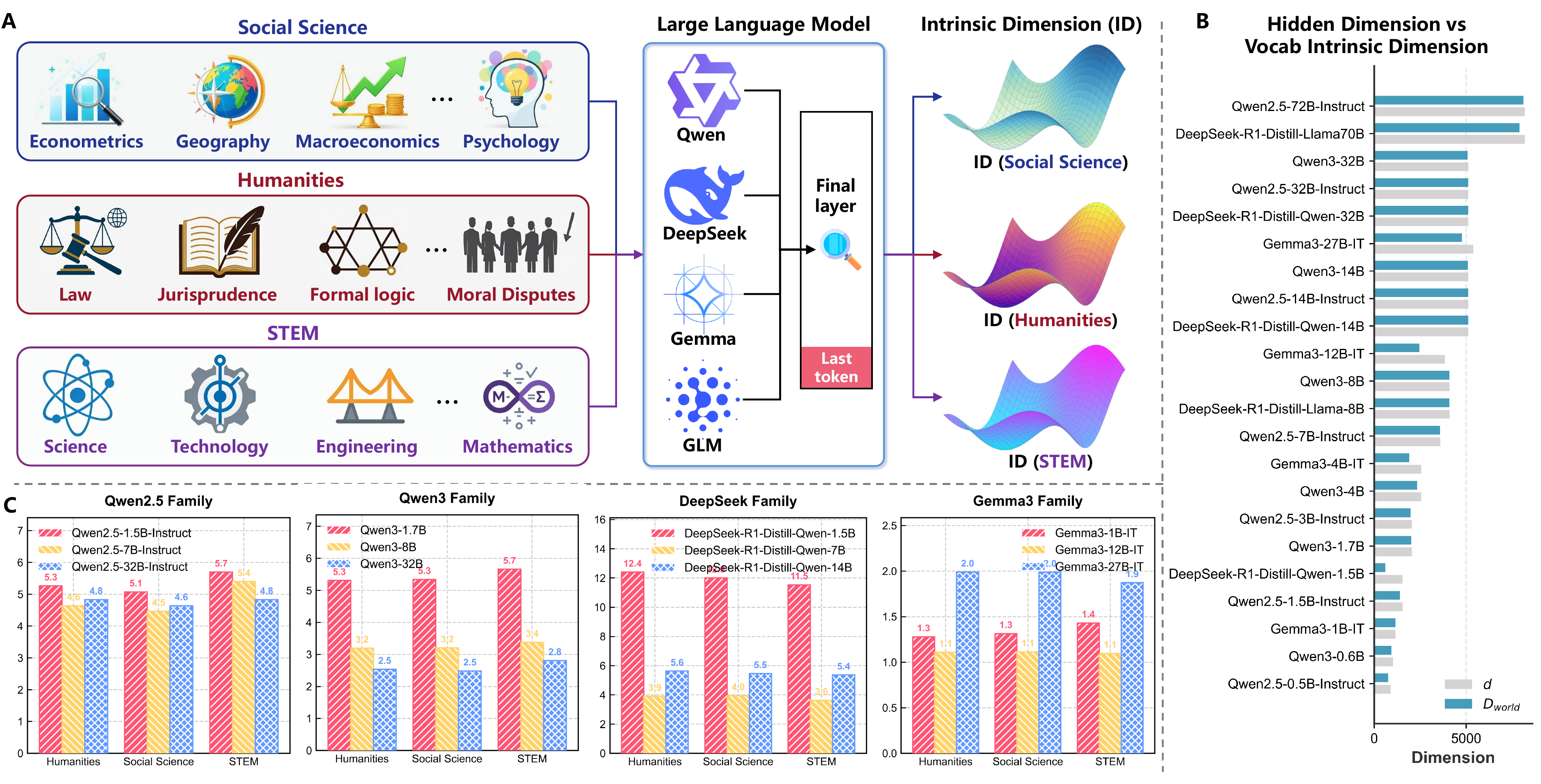}
\caption{\textbf{Low-dimensional organization is robust across stimuli and decoupled from global representational capacity.} \textbf{A,} Distribution of stimulus-induced intrinsic dimensionality (ID; Two-Nearest-Neighbor local estimator, TLE) across heterogeneous generic cognitive stimuli, demonstrating that inference-time trajectories consistently concentrate on low-dimensional manifolds across prompts and domains. \textbf{B,} Intrinsic dimensionality of static vocabulary embeddings for the same model set, which remains near the ambient embedding dimension, indicating that the representational substrate retains high expressivity and that dimensional collapse is a stimulus-dependent property of inference dynamics rather than a global bottleneck.}
\label{fig2}
\end{figure*}

\section*{Results}

\subsection*{Inference-time reasoning dynamics self-organize into low-dimensional manifolds}

To characterize reasoning as an internal process rather than a simple input–output mapping, we analyze inference-time representation trajectories elicited by generic cognitive stimuli \cite{wang2024mmlu} across a range of contemporary large language models \cite{bai2025qwen25vltechnicalreport,yang2025qwen3,gemmateam2025gemma3technicalreport,guo2025deepseek}. For each stimulus, at each layer and at each inference step, we record the hidden state of the last token in the current sequence. The sequence of such states within each layer is treated as a discrete trajectory embedded in the model’s high-dimensional representation space.

Across all evaluated models, \uline{we observe a striking and highly reproducible phenomenon:} despite being embedded in representation spaces with thousands of dimensions, reasoning trajectories consistently and spontaneously collapse onto low-dimensional manifolds during inference.
As shown in \textcolor{red}{Figure \ref{fig1}}, the intrinsic dimensionality \cite{ansuini2019intrinsic,ma2024unveiling} of stimulus-induced representations decreases rapidly as inference proceeds through network layers and stabilizes at values far below the ambient embedding dimension. In many cases, the intrinsic dimensionality converges to fewer than ten degrees of freedom. We further observe differences across model: weaker or earlier models tend to exhibit slower and less stable dimensional reduction, whereas newer-generation models converge more consistently to compact manifolds.

This dimensional collapse is robust across model families and parameter scales and is not confined to specific prompts or task types. As illustrated in \textcolor{red}{Figure \ref{fig2}}A, low-dimensional organization emerges consistently under heterogeneous cognitive stimuli, including commonsense reasoning, linguistic reasoning, and conceptual inference. Importantly, this structure arises dynamically during inference rather than being imposed by architectural bottlenecks or explicit regularization, indicating that low-dimensional organization is a spontaneous and general property of reasoning dynamics in modern language models. The emergence of compact reasoning manifolds stands in sharp contrast to the expressive capacity of the underlying representation space itself.

To distinguish stimulus-induced organization from global representational limitations, we estimate the intrinsic dimensionality of general-purpose representations using static vocabulary embeddings. As shown in \textcolor{red}{Figure \ref{fig2}}B, across all models these embeddings occupy spaces whose intrinsic dimensionality remains close to the ambient dimension, indicating that the models retain a high-capacity substrate for representing diverse world concepts. The coexistence of near–full-dimensional world representations and sharply compressed reasoning trajectories demonstrates that dimensional collapse during reasoning does not reflect representational degradation, but rather a stimulus-dependent specialization of inference dynamics.

\uline{Together, these results establish a fundamental empirical property of large language models:} reasoning is realized as a low-dimensional dynamical process embedded within a highly expressive representation space. Inference-time dynamics are neither diffuse explorations of the ambient space nor static mappings between isolated representations. Instead, they self-organize into compact manifolds that constrain the evolution of internal states during reasoning. At the same time, dimensional collapse alone is insufficient to distinguish robust reasoning from brittle or degenerate behavior. As we show in the following section, models with similarly low-dimensional reasoning trajectories can exhibit qualitatively different inference dynamics, indicating that additional structural constraints are required to characterize healthy reasoning processes.

\begin{figure*}[!ht]
\centering
\includegraphics[width=0.98\linewidth]{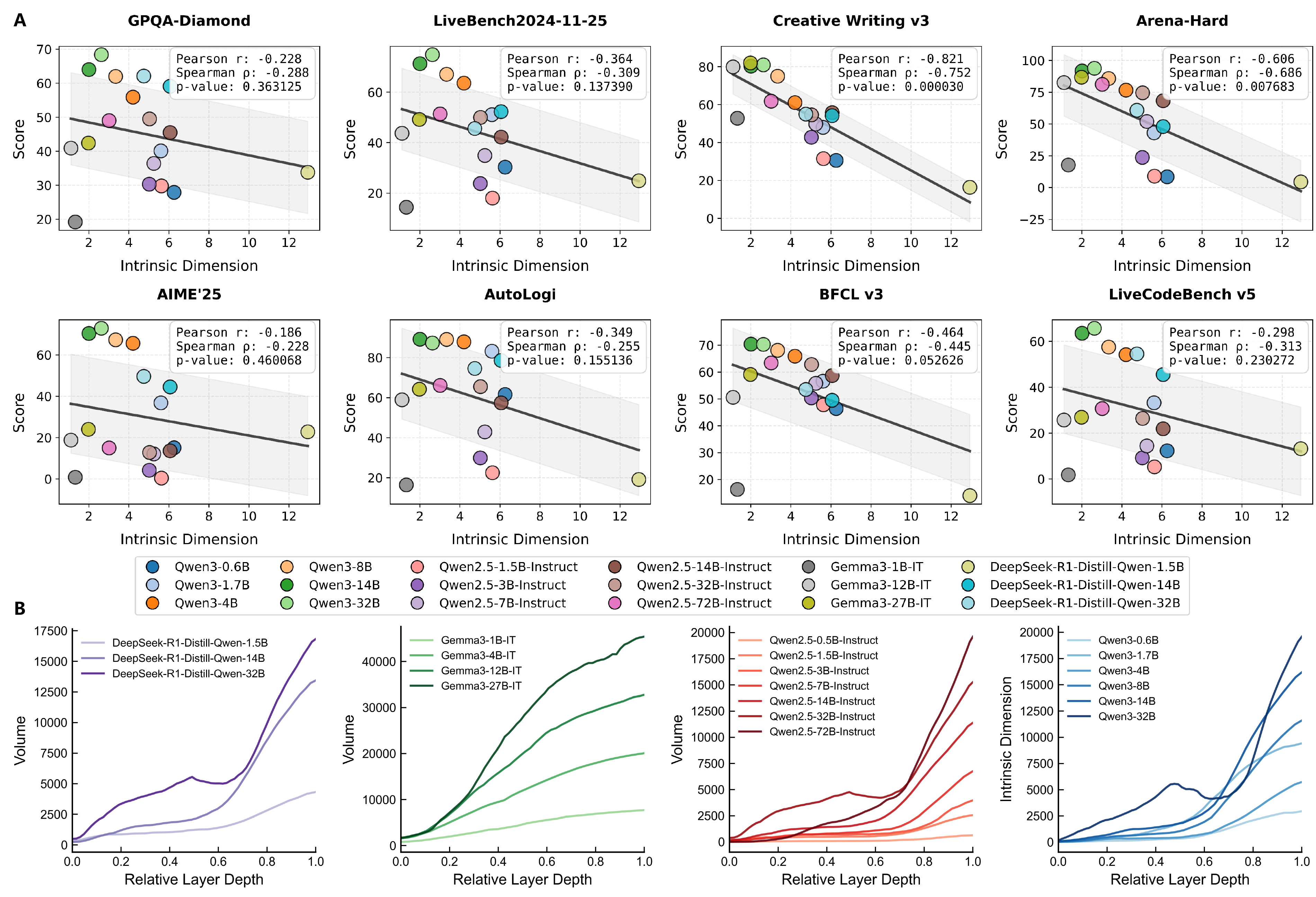}
\caption{\textbf{Geometric compression alone does not explain robust reasoning.} 
\textbf{A,} Relationship between stimulus-induced intrinsic dimensionality (ID) of inference-time representation trajectories and downstream reasoning performance across models of varying families and scales, showing that stronger compression (lower ID) does not consistently correspond to better performance. 
\textbf{B,} Joint evolution of geometry and information across depth: as inference progresses through layers, trajectory ID decreases while information-carrying capacity (information volume $V$; see Methods) increases, indicating that effective reasoning requires low-dimensional organization together with non-degenerate information flow.}
\label{fig3}
\end{figure*}

\begin{figure*}[!ht]
\centering
\includegraphics[width=0.98\linewidth]{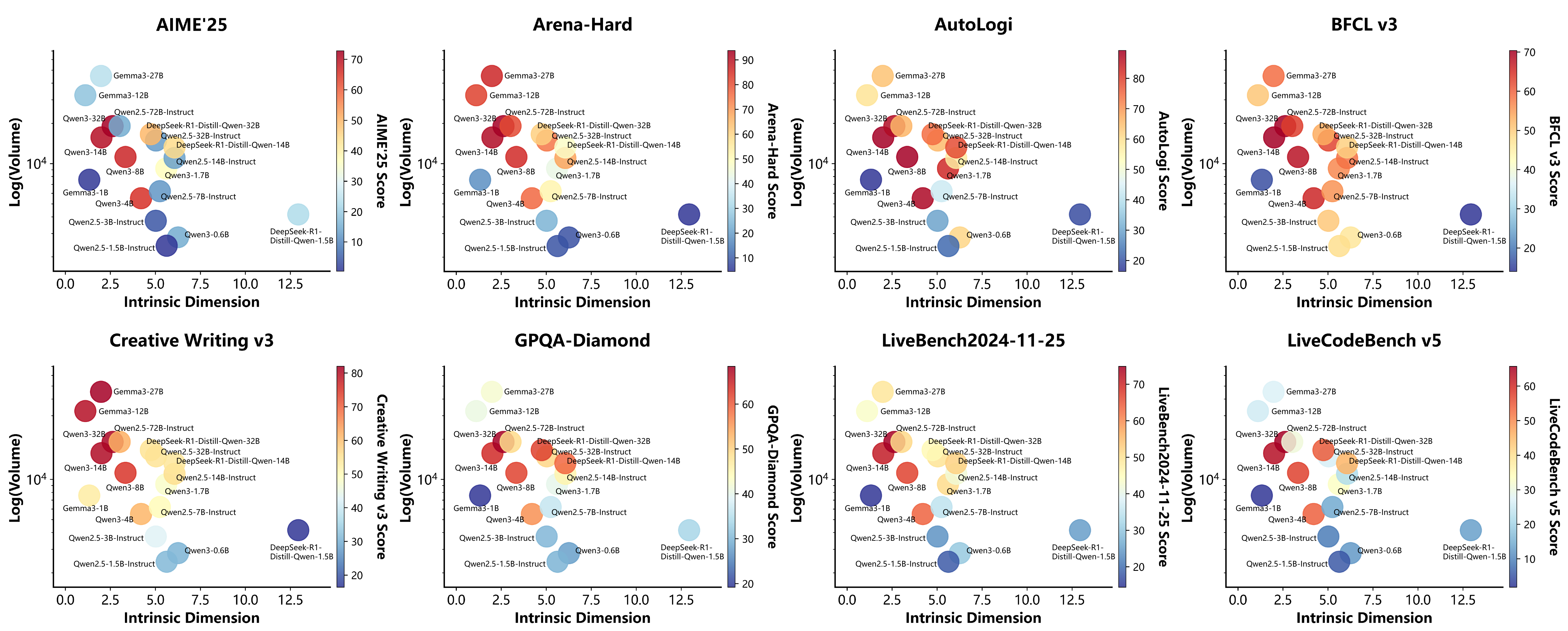}
\caption{\textbf{An admissible regime of inference dynamics jointly constrains geometry, information and performance.} Models are positioned in a three-dimensional space defined by stimulus-induced intrinsic dimensionality (ID), information-carrying capacity of inference trajectories (information volume $V$; Methods) and downstream reasoning performance. High-performing models concentrate within a narrow region characterized by low ID and elevated $V$, whereas models outside this regime exhibit degraded or unstable performance, consistent with the proposed structural constraints on healthy reasoning.}
\label{fig4}
\end{figure*}

\subsection*{Low-dimensional structure alone does not ensure robust reasoning}

The spontaneous collapse of inference-time trajectories onto low-dimensional manifolds naturally invites a simple interpretation: that stronger reasoning corresponds to stronger compression, and that progressively lower intrinsic dimensionality implies healthier inference dynamics. From a theoretical perspective, however, dimensionality reduction alone is insufficient to guarantee meaningful computation. In the extreme, unconstrained compression would be expected to drive reasoning trajectories toward rigid low-dimensional curves or near-fixed points, eliminating the degrees of freedom required to represent intermediate computation.

Empirically, we find that intrinsic dimensionality by itself does not reliably predict reasoning performance. Across multiple model families and parameter scales, models whose stimulus-induced trajectories exhibit lower intrinsic dimensionality do not consistently outperform those operating at slightly higher dimensionality. In the representative examples shown in \textcolor{red}{Figure \ref{fig3}}A, more aggressively compressed models achieve reasoning performance comparable to, or even worse than, models with weaker compression. These observations indicate that, although low dimensionality is necessary for structured reasoning, it is insufficient to characterize reasoning quality.

This limitation motivates the introduction of an additional quantity that captures how much information is actively present within the reasoning manifold. To this end, we examine the joint evolution of geometric compression and information content across network depth. We observe that, as inference proceeds through deeper layers, the intrinsic dimensionality of reasoning trajectories decreases systematically. In contrast, the information content of the reasoning manifold, quantified by the information volume defined in Methods, exhibits a markedly different trend: across all evaluated models, it increases with depth (\textcolor{red}{Figure \ref{fig3}}B).

\uline{This result demonstrates that dimensional compression during reasoning does not imply information loss.} Instead, it reflects a focusing mechanism: deeper layers suppress irrelevant noise (reducing dimensionality) while amplifying task-relevant conceptual variations (increasing information volume).
Early layers typically exhibit higher-dimensional but information-sparse dynamics, whereas deeper layers operate on highly constrained manifolds that nonetheless sustain substantially greater information content.
\uline{These findings indicate that effective reasoning relies on maintaining high information content within a compact geometric structure, rather than on indiscriminate compression.}

Crucially, intrinsic dimensionality alone cannot distinguish between these two mechanisms. Two models may exhibit similar degrees of geometric compression while differing substantially in how much information is preserved and organized within their reasoning trajectories. The observed increase in information content with depth provides direct empirical evidence that healthy reasoning dynamics do not trade off geometric constraint against information preservation, but instead integrate the two in a coordinated manner.

\textcolor{red}{Figure \ref{fig4}} visualizes this relationship by jointly plotting intrinsic dimensionality, information-carrying capacity, and reasoning performance across multiple model families. In this three-dimensional landscape, each model is positioned according to its stimulus-induced intrinsic dimensionality, the information content of its reasoning manifold, and downstream reasoning performance. Across architectures and scales, high-performing models are not distributed along a single axis, but instead concentrate within a confined region characterized by relatively low dimensionality and elevated information content. Models located at either extreme, corresponding to excessively compressed or insufficiently informative regimes, tend to exhibit degraded or unstable performance.

Taken together, these results establish that \uline{intrinsic dimensionality alone cannot characterize healthy reasoning dynamics. Geometric compression is a necessary component of structured inference, but it must be accompanied by sufficient information preservation to support meaningful internal computation.} 

\begin{figure*}[!ht]
\centering
\includegraphics[width=1\linewidth]{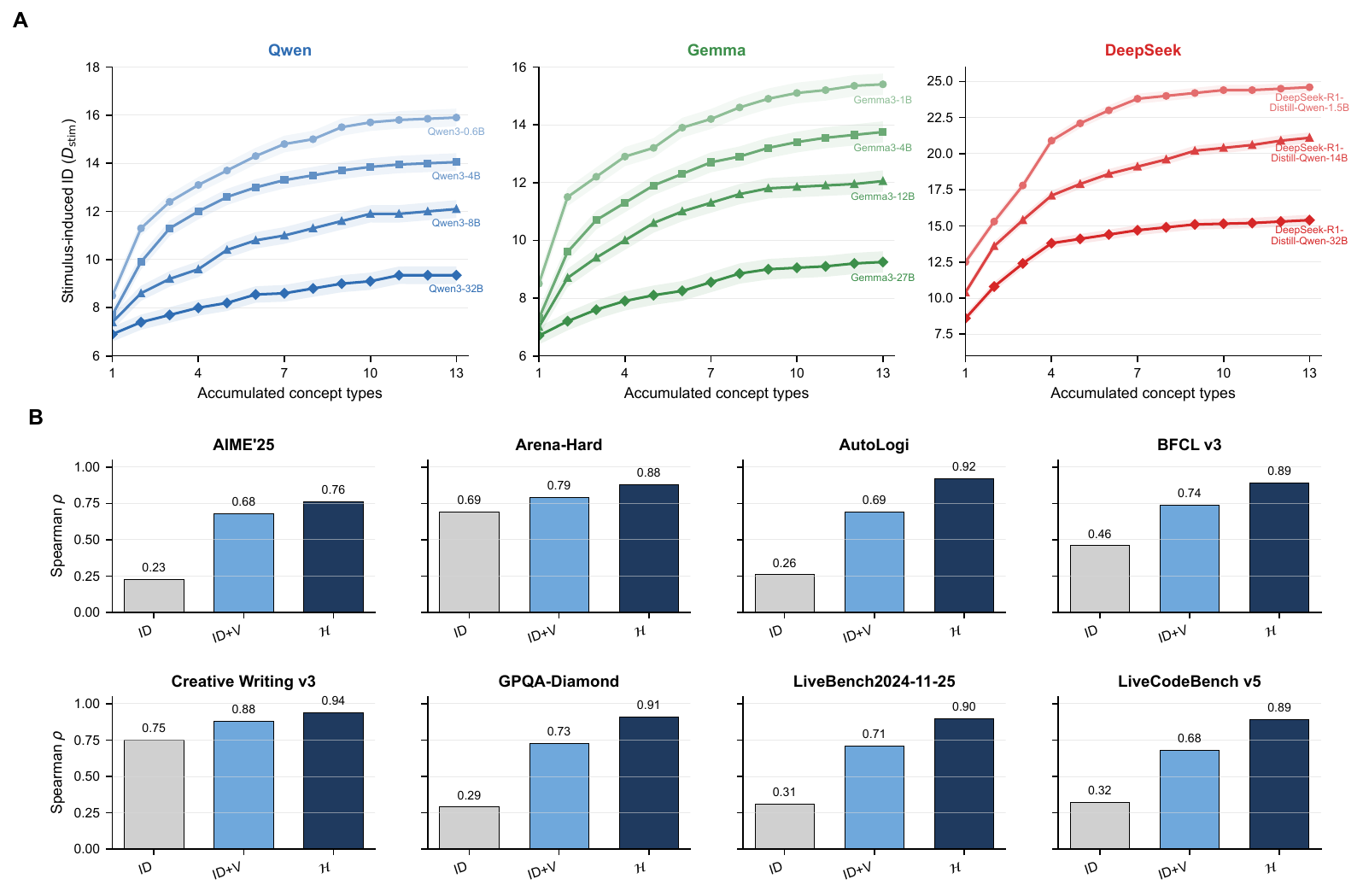}
\caption{
\textbf{Expressive capacity stabilizes inference dynamics and enables a unified structural diagnostic.}
\textbf{A,} Stimulus expansion analysis showing that stimulus-induced intrinsic dimensionality increases with accumulated conceptual diversity, with higher-expressivity models exhibiting slower expansion.
\textbf{B,} Benchmark-wise ablation of structural predictors. For each benchmark, bars show the Spearman correlation between model performance and intrinsic dimensionality alone, intrinsic dimensionality combined with information volume, and the full diagnostic $\mathcal H$. The full diagnostic consistently provides the strongest association with benchmark performance.
}
\label{fig5}
\end{figure*}

\subsection*{Expressive capacity stabilizes low-dimensional reasoning under increasing conceptual diversity}

The results in the preceding sections establish that low-dimensional organization and sufficient information content are necessary components of healthy reasoning dynamics. However, even when both conditions are satisfied, inference trajectories can become unstable as the conceptual diversity of input stimuli increases. This suggests the presence of an additional structural factor governing the robustness of low-dimensional reasoning under heterogeneous inputs.
We refer to this factor as the model’s expressive capacity, defined as the intrinsic dimensionality of its static vocabulary\footnote{In transformer-based language models, the static vocabulary embedding matrix maps discrete tokens to continuous vectors prior to contextual processing. It defines the representational substrate of the model, determining the degrees of freedom available for encoding lexical and semantic concepts independently of any specific input or reasoning trajectory.} embedding space. This quantity captures the effective degrees of freedom available for representing diverse world concepts, independently of any specific reasoning task. Importantly, expressive capacity characterizes the representational substrate on which inference-time dynamics operate, rather than the dynamics themselves.

To examine how expressive capacity influences the stability of reasoning manifolds, we design a controlled stimulus expansion experiment that progressively increases conceptual diversity while tracking changes in stimulus-induced intrinsic dimensionality. We partition the MMLU-Other benchmark into thirteen disjoint question types, each probing a distinct category of conceptual or cognitive content. For each model, we first estimate inference-time intrinsic dimensionality using stimuli drawn from a single question type. We then incrementally expand the stimulus set by cumulatively adding additional question types and recompute the stimulus-induced intrinsic dimensionality at each stage.

This procedure yields a growth curve that quantifies how the effective dimensionality of reasoning trajectories responds to increasing conceptual diversity. Across models, we observe a clear and systematic pattern: models with greater expressive capacity exhibit significantly slower growth of stimulus-induced intrinsic dimensionality as conceptual diversity increases (\textcolor{red}{Figure \ref{fig5}}A). In contrast, models with more limited expressive capacity show a more rapid expansion of reasoning dimensionality when confronted with heterogeneous stimuli.

\uline{These results indicate that expressive capacity stabilizes low-dimensional reasoning organization under stimulus expansion.} Models with richer representational substrates are able to accommodate increasing conceptual diversity without requiring substantial expansion of inference-time degrees of freedom. By contrast, models with insufficient expressive capacity must recruit additional dimensions during inference to encode heterogeneous concepts, leading to dispersion of reasoning trajectories and reduced structural stability.
From this perspective, expressive capacity does not directly determine reasoning performance, nor does it operate as a proxy for model size or task accuracy. Instead, it functions as a background structural condition that modulates how effectively low-dimensional, information-rich reasoning dynamics can be sustained as input diversity increases. Expressive capacity is therefore neither inherently beneficial nor detrimental; rather, it conditions the regime within which inference dynamics can remain compact and coordinated.

In summary, these results establish expressive capacity as a necessary structural factor for robust reasoning dynamics. Even when inference trajectories are geometrically compact and information-rich, their stability under heterogeneous stimuli depends on the representational capacity of the underlying embedding space. In the next section, we show how expressive capacity, together with geometric organization and information content, can be integrated into a unified, label-free characterization of reasoning health.

\subsection*{A unified structural diagnostic of reasoning}

The preceding sections identify three independent structural constraints governing healthy reasoning dynamics: low-dimensional geometric organization, sufficient information content within the reasoning manifold, and sufficient expressive capacity of the underlying representation space. Together, these findings motivate a unified, label-free characterization of reasoning health that operates directly on inference-time representations.

We formalize this characterization through a scalar diagnostic that integrates the three structural components into a single quantity. Specifically, for each model we define a reasoning health score
\begin{equation}
\mathcal{H}
=
\log(D_{\mathrm{world}})
\cdot
\frac{V}{\exp(\varepsilon D_{\mathrm{stim}})},
\end{equation}
where $D_{\mathrm{world}}$ denotes the intrinsic dimensionality of static vocabulary embeddings, capturing expressive capacity; $D_{\mathrm{stim}}$ denotes the stimulus-induced intrinsic dimensionality of inference-time trajectories, capturing geometric organization; and $V$ denotes the information-carrying capacity of the reasoning manifold. The parameter $\varepsilon$ controls the strength of the penalty on diffuse inference dynamics and is fixed across all models. All quantities are computed solely from internal representations, as detailed in Methods.

This formulation reflects the structural constraints revealed by our empirical analyses. The numerator rewards inference dynamics that sustain substantial information content, while the exponential penalty suppresses regimes in which reasoning trajectories expand excessively in dimensionality. The logarithmic dependence on expressive capacity reflects its role as a background condition that supports stable organization without dominating the diagnostic. Importantly, the diagnostic does not incorporate task labels, correctness signals, or external supervision, and is not used for training or model selection.

To assess whether this structural diagnostic captures meaningful variation in reasoning behavior, we compare the resulting scores against model performance on a range of independent reasoning benchmarks. For each model, we compute $\mathcal{H}$ from generic cognitive stimuli and evaluate its rank correlation with benchmark scores reported on external leaderboards. Across all evaluated benchmarks, we observe consistently strong monotonic agreement (Spearman rank correlation coefficients exceeding 0.9 across all evaluated benchmarks; \textcolor{red}{Figure} \ref{fig5}B).

Crucially, this alignment emerges despite the diagnostic being entirely task-agnostic. The computation of $\mathcal{H}$ does not reference benchmark tasks, evaluation datasets, or performance labels, and relies only on internal inference dynamics elicited by generic stimuli. This result indicates that models exhibiting structurally healthy reasoning dynamics, as characterized by geometric constraint, information content, and expressive support, tend to generalize robustly across diverse reasoning tasks.

At the same time, the diagnostic is not intended to replace benchmark-based evaluation. Rather, it provides a complementary lens that isolates intrinsic properties of inference dynamics from dataset-specific effects, prompt sensitivity, and surface-level correctness. Models with similar benchmark performance can occupy distinct regions of the intrinsic constraint space, while models with structurally healthy dynamics may underperform on specific tasks due to misalignment or insufficient supervision.

Together, these results demonstrate that the structural constraints identified in this work can be operationalized into a unified, label-free diagnostic that captures intrinsic aspects of reasoning health. By grounding reasoning evaluation in the geometry and information flow of inference-time dynamics, this approach offers a principled means of comparing, diagnosing, and understanding reasoning behavior across large language models, independent of task-specific supervision.

\section*{Discussion}

The results presented in this work motivate a shift in how reasoning in large language models is conceptualized and evaluated. Rather than focusing exclusively on task-level outcomes or benchmark scores, we examine reasoning through the internal structure of inference-time dynamics. This perspective treats reasoning as a constrained dynamical process unfolding within a representational space~\cite{joshi2025geometry}, and invites questions not only about whether a model succeeds on a task, but about the internal regimes that make such success possible or fragile. The Discussion below situates our findings within this broader conceptual framework, clarifies what is captured by structural measures of reasoning health, delineates the regime in which the framework applies, and outlines how such structural diagnostics may be used in practice.


\subsection*{What is being measured by reasoning health?}

A central contribution of this work is to clarify what is captured by internal measures of reasoning quality, as well as what is not. Rather than estimating task-specific competence or benchmark performance, the reasoning health diagnostic introduced here quantifies the \emph{structural organization of inference-time dynamics}. It evaluates whether internal representations evolve within a regime that supports meaningful intermediate computation, independent of whether a particular output is correct~\cite{turpin2023language,peng2025sampling}.

From this perspective, reasoning health characterizes how a model reasons, not what it knows or how well it performs on a given dataset. Models with similar task accuracy may operate in fundamentally different internal regimes \cite{tanneru2024hardness,guo2025deepseek}, while models with structurally healthy inference dynamics may fail specific benchmarks due to misalignment, insufficient supervision, or domain mismatch. The diagnostic therefore complements, rather than replaces, external evaluation \cite{liang2022holistic} by isolating intrinsic properties of inference dynamics that are otherwise conflated with dataset effects. In this sense, reasoning health provides a measure of whether a system operates within a regime capable of sustaining intermediate computational structure, rather than merely producing correct outputs.

\subsection*{Structural determinants of healthy reasoning dynamics}

Our results identify three structural factors that jointly govern reasoning dynamics: low-dimensional manifold structure, information content, and expressive capacity. Low-dimensional manifold structure constrains inference trajectories into coordinated internal subspaces, preventing diffuse and unstable exploration. Information content ensures that these manifolds retain sufficient structured variation to support intermediate computation rather than collapsing into trivial dynamics. Expressive capacity, quantified independently through static embedding geometry, provides the representational substrate that stabilizes low-dimensional reasoning manifolds under increasing conceptual diversity.

Crucially, none of these factors alone is sufficient. Dimensional compression without sufficient information content leads to rigid, uninformative dynamics, while information-rich representations without low-dimensional geometric constraint remain unstable. Likewise, expressive capacity conditions how inference dynamics respond to heterogeneous stimuli, enabling models to maintain compact reasoning trajectories as conceptual diversity increases. Healthy reasoning thus emerges not from any single mechanism, but from the coordinated interaction of these structural determinants.

\subsection*{Regime of validity of the admissible reasoning framework}

The framework developed in this work applies specifically to inference-time reasoning dynamics in contemporary autoregressive language models. It characterizes a regime in which reasoning unfolds as a low-dimensional, information-rich dynamical process embedded within a highly expressive representation space. Outside this regime, inference dynamics exhibit characteristic failure modes, including diffuse exploration, information starvation, or instability under stimulus expansion.

Importantly, the admissible reasoning regime is not expected to extend indefinitely with model scale, architectural complexity, or training compute~\cite{bai2025qwen25vltechnicalreport}. Changes that improve one structural dimension may degrade another, shifting models toward or away from the admissible regime. The framework therefore does not predict monotonic scaling laws for reasoning performance~\cite{xiao2025densing}, but instead delineates a \emph{structural window} within which robust reasoning dynamics can be sustained. Whether similar regimes govern reasoning in non-autoregressive architectures, multimodal systems, or embodied agents remains an open question.

\subsection*{Applications}

By grounding reasoning evaluation in internal dynamics rather than task outcomes, the diagnostic introduced here enables comparison of reasoning behavior across models, architectures, and training regimes without reliance on labeled benchmarks. This makes it applicable in settings where task-specific evaluation is limited or unavailable.

More broadly, structural diagnostics provide a complementary perspective for monitoring and guiding model development. Instead of optimizing solely for benchmark performance, training or fine-tuning procedures may be informed by constraints on inference dynamics~\cite{snell2024scaling}, promoting regimes that support robust and generalizable reasoning. Structural analysis may also help identify brittleness, characterize failure modes, and assess how alignment or compression techniques affect internal computation.

From this perspective, the identification of an admissible structural regime raises questions about how such regimes might be encouraged during model development. One possible implication, not examined here, is whether training objectives that bias inference-time dynamics toward geometrically admissible regimes could promote healthier reasoning behavior. This does not imply a specific regularization strategy. Rather, the admissible regime provides a structural reference for evaluating the effects of future training approaches.

Finally, this framework suggests a shift in how reasoning in large artificial systems is studied. By treating reasoning as a constrained dynamical process with identifiable structural regimes, it becomes possible to analyze intelligence at a level that is more general and more mechanistic than task-specific evaluation.

\subsection*{Limitations and scope}
Several caveats delimit the scope of our claims. First, last-token hidden states provide a tractable proxy for inference dynamics but do not exhaust computation distributed across positions, heads, or attention patterns. Second, the stimuli used here are label-free but not distribution-free, as they are derived from benchmark-style prompts rather than unconstrained naturalistic interaction. Third, the proposed diagnostic is descriptive and correlational: it identifies a structural regime associated with strong reasoning behavior, but does not by itself establish causal mechanisms. Finally, whether similar admissible regimes govern non-autoregressive, multimodal, or embodied systems remains an open question.

\section*{Methods}

\subsection*{Models and inference configuration}

We evaluated contemporary open-weight autoregressive large language models from four model families: Qwen2.5, Qwen3, Gemma 3, and DeepSeek-R1-Distill-Qwen. The Qwen2.5 series spanned parameter scales from 0.5B to 72B, the Qwen3 series spanned 0.6B to 32B, Gemma 3 included the 1B, 4B, 12B, and 27B checkpoints, and DeepSeek-R1-Distill-Qwen included the 1.5B, 14B, and 32B checkpoints. All model weights were publicly released and were used in inference-only mode, without additional fine-tuning or parameter updates.

Unless otherwise stated, all models were evaluated with the same prompt wrapper and the same decoding configuration. Responses were sampled at temperature \(0.7\), and the maximum generation length was set to \(15{,}000\) tokens. Each prompt generated a single completion per run. No task-specific prompt optimization, self-consistency voting, answer extraction heuristics, or post-hoc filtering was applied. All internal metrics were computed solely from hidden-state dynamics and did not access answer keys, correctness labels, or benchmark scores.

\subsection*{Stimulus construction}

To probe reasoning dynamics independently of task supervision, we constructed a set of generic cognitive stimuli from the ``Other'' subset of the Massive Multitask Language Understanding (MMLU) benchmark. This subset spans a broad range of conceptual, linguistic, and commonsense questions while avoiding reliance on any single narrow reasoning domain.

Stimuli were used only to elicit inference-time dynamics. No answer labels were loaded during the computation of internal metrics, and no notion of correctness entered the geometric or information-theoretic analyses. This design ensured that all reported structural quantities reflected intrinsic properties of inference dynamics rather than benchmark alignment.
For the stimulus-expansion analysis, the MMLU-Other prompts were partitioned into thirteen disjoint question types defined from the source benchmark. Progressively larger stimulus sets were then constructed by cumulatively adding question types and recomputing the corresponding structural quantities at each stage.

\subsection*{Inference-time representation extraction}

For each stimulus \(x\), models generated responses autoregressively until termination by EOS or the maximum generation length of 15{,}000 tokens. At inference step \(t\) and transformer layer \(\ell\), we extracted the hidden state at the final sequence position, denoted
\begin{equation}
h_{\ell,t}(x) \in \mathbb{R}^{d_\ell},
\end{equation}
where \(d_\ell\) is the hidden width of layer \(\ell\).

A layer-specific reasoning trajectory was defined as the ordered sequence
\begin{equation}
\mathcal{T}_{\ell}(x)=\{h_{\ell,1}(x),h_{\ell,2}(x),\dots,h_{\ell,T(x)}(x)\},
\end{equation}
where \(T(x)\) denotes the number of generated tokens for stimulus \(x\). Unless otherwise stated, all analyses were performed on the final-token hidden representations at each transformer layer. For cross-model summary statistics, stimulus-induced intrinsic dimensionality and information volume were aggregated across prompts at the final layer.

\subsection*{Preprocessing of representations}

For each prompt and layer, hidden states were mean-centred before geometric analysis. Specifically,
\begin{equation}
\begin{split}
\tilde{h}_{\ell,t}(x)=h_{\ell,t}(x)-\bar{h}_{\ell}(x), \\
\bar{h}_{\ell}(x)=\frac{1}{T(x)}\sum_{t=1}^{T(x)} h_{\ell,t}(x).
\end{split}
\end{equation}

All intrinsic-dimensionality and information-volume calculations were performed on centred representations. Unless otherwise stated, distances were computed in Euclidean geometry without additional dimensionality reduction.

\subsection*{Intrinsic dimensionality estimation}

Intrinsic dimensionality (ID) quantifies the effective number of degrees of freedom governing representations concentrated near a low-dimensional manifold embedded in a high-dimensional space \cite{durstewitz2025neuroscience,ma2024unveiling}. Given a set of points
\begin{equation}
Z=\{z_i\}_{i=1}^{m}\subset\mathbb{R}^{d},
\end{equation}
we estimated intrinsic dimensionality using the tight local intrinsic dimensionality estimator (TLE) \cite{ma2024unveiling,facco2017estimating}, a \(k\)-nearest-neighbour estimator designed for locally sampled manifolds in high-dimensional spaces.

For each point \(z_i\), let \(V_i\) denote its \(k\)-nearest neighbours, let \(V_i^\ast = V_i \cup \{z_i\}\), and let \(r_k(z_i)\) denote the distance from \(z_i\) to its \(k\)-th nearest neighbour. The local intrinsic dimensionality at \(z_i\) was computed as
\begin{equation}
\begin{split}
ID_{\mathrm{TLE}}(z_i)
= -\Bigg(
\frac{1}{|V_i^\ast|^2}
\sum_{\substack{v,w\in V_i^\ast\\v\neq w}}
\bigg[
& \log\frac{d_{z_i}(v,w)}{r_k(z_i)} \\
& + \log\frac{d_{z_i}(2z_i-v,w)}{r_k(z_i)}
\bigg]
\Bigg)^{-1},
\end{split}
\end{equation}
where
\begin{equation}
d_{z_i}(v,w)=
\frac{r_k(z_i)\,(w-v)^\top(w-v)}
{2\,(z_i-v)^\top(w-v)}.
\end{equation}

The global intrinsic dimensionality of \(Z\) was obtained by averaging local estimates across points:
\begin{equation}
ID_{\mathrm{TLE}}(Z)
=
\frac{1}{m}
\sum_{i=1}^{m} ID_{\mathrm{TLE}}(z_i).
\end{equation}

We estimated two conceptually distinct intrinsic dimensionalities. First, the world expressivity
\begin{equation}
D_{\mathrm{world}},
\end{equation}
was computed from the static vocabulary embedding matrix of each model. Second, the stimulus-induced dimensionality
\begin{equation}
D_{\mathrm{stim}},
\end{equation}
was computed from inference-time trajectories elicited by the generic cognitive stimuli. At layer \(\ell\), the per-prompt quantity was
\begin{equation}
D_{\mathrm{stim},\ell}(x)=ID_{\mathrm{TLE}}(\mathcal{T}_{\ell}(x)),
\end{equation}
and the model-level summary was obtained by aggregating across prompts at the final layer. Unless otherwise stated, we used \(k=20\) neighbours and Euclidean distance throughout.

\subsection*{Information volume of inference trajectories}

To quantify the amount of structured variation preserved during inference, we computed the information volume of each trajectory. For a given prompt \(x\) and layer \(\ell\), we first defined the trajectory mean
\begin{equation}
\bar{h}_{\ell}(x)=\frac{1}{T(x)}\sum_{t=1}^{T(x)} h_{\ell,t}(x),
\end{equation}
and the centred trajectory matrix
\begin{equation}
\begin{split}
Z_{\ell}(x)=
\bigl[
h_{\ell,1}(x)-\bar{h}_{\ell}(x),\;
h_{\ell,2}(x)-\bar{h}_{\ell}(x), \\
\dots,\;
h_{\ell,T(x)}(x)-\bar{h}_{\ell}(x)
\bigr]
\in\mathbb{R}^{d_\ell\times T(x)}.
\end{split}
\end{equation}

The information volume \cite{ma2025predicting,wright2010sparse} was then defined as
\begin{equation}
V_{\ell}(x)
=
\frac{1}{2}
\log\det\left(
I+\frac{d_\ell}{T(x)}Z_{\ell}(x)Z_{\ell}(x)^\top
\right).
\end{equation}

This quantity is small for trajectories that collapse toward fixed points or near-degenerate curves, and larger for compact trajectories that retain substantial spread along informative directions. Model-level summaries of information volume were obtained by aggregating \(V_{\ell}(x)\) across prompts at the final layer.

\subsection*{Label-free reasoning health diagnostic}

To integrate the structural constraints identified in the main text, we defined a unified, label-free diagnostic of reasoning dynamics:
\begin{equation}
\mathcal{H}
=
\log(D_{\mathrm{world}})
\cdot
\frac{V}{\exp(\varepsilon D_{\mathrm{stim}})},
\end{equation}
where \(D_{\mathrm{world}}\) denotes the intrinsic dimensionality of static vocabulary embeddings, capturing representational expressivity; \(D_{\mathrm{stim}}\) denotes the stimulus-induced intrinsic dimensionality of inference-time trajectories, capturing geometric organization; and \(V\) denotes the information volume of those trajectories, capturing the amount of structured variation preserved within the compressed manifold.

The penalty coefficient \(\varepsilon\) was fixed \emph{a priori} to \(0.1\) for all models. This value was chosen to penalize diffuse inference dynamics without allowing the dimensionality term to dominate the contributions of information volume and representational expressivity. The value of \(\varepsilon\) was not tuned separately for any benchmark or model family.

All three quantities were computed solely from internal representations. The diagnostic did not access task labels, reference answers, or benchmark correctness signals, and was used exclusively for post-hoc structural analysis rather than model training or selection.

\subsection*{Control analyses}

To verify that the observed low-dimensional organization reflected structured inference rather than superficial properties of autoregressive generation, we performed several control analyses. First, token order within each trajectory was randomly shuffled before recomputing intrinsic dimensionality and information volume. Second, the generic cognitive stimuli were replaced with non-cognitive or weakly structured prompts while maintaining the same decoding configuration (temperature \(=0.7\); maximum generation length \(=15{,}000\) tokens). Third, we recomputed all metrics on truncated trajectories to confirm that the reported geometry was not driven by unusually long generations.

These controls did not reproduce the stable, compact manifolds observed under generic cognitive stimulation, indicating that the reported low-dimensional organization was specifically associated with structured inference dynamics.

\subsection*{Statistical analysis and reproducibility}

No statistical method was used to predetermine sample size. Sample sizes were determined by the number of models and stimuli included in each analysis.

Unless otherwise stated, uncertainty intervals are reported as two-sided 95\% bootstrap confidence intervals estimated from \(N_{\mathrm{boot}}=1000\) resamples. Bootstrap resampling was performed with replacement over stimuli. For the stimulus-expansion analysis, resampling was performed within question types before aggregation across cumulatively expanded stimulus sets. For benchmark-correlation analyses, models were resampled with replacement.

Because all models were evaluated under the same decoding temperature (\(0.7\)) and the same maximum generation length (\(15{,}000\) tokens), cross-model comparisons were not confounded by differences in generation hyperparameters. Qualitative trends reported in the Results were consistent across model families, parameter scales, and stimulus subsets.

\subsection*{Implementation details}

All experiments were conducted using publicly released model weights in inference-only mode with standard open-source deep-learning libraries. Hidden states were extracted during autoregressive generation and stored for post-hoc geometric analysis. Nearest-neighbour computations for intrinsic-dimensionality estimation and linear-algebra operations for information-volume computation were performed using GPU-accelerated routines where available.

All analysis code was deterministic conditioned on fixed model weights, prompts, generated trajectories, and random seeds. Because response generation was performed with sampling temperature set to \(0.7\), inference trajectories were stochastic at generation time unless random seeds were fixed explicitly.

\section*{Data availability}


The prompt texts analysed in this study were derived from the public MMLU benchmark, subject to its original licence and terms of use. The source data underlying Figs.~\ref{fig1}--\ref{fig5}, together with the per-model summary statistics used for all geometric, information-theoretic, and correlation analyses, will be deposited in a public repository upon publication. Public benchmark scores used for the correlation analyses were obtained from the corresponding public evaluation sources and will be listed in the Supplementary Materials together with access dates and checkpoint matching details.

\section*{Code availability}


Custom code for hidden-state extraction, intrinsic-dimensionality estimation, information-volume computation, stimulus-expansion analysis, the structural diagnostic \(\mathcal{H}\), and figure generation will be made publicly available upon publication. Intrinsic-dimensionality estimation leverages the \texttt{perceptual-manifold-geometry} package~\citep{ma2024unveiling}, which provides geometric analysis tools for high-dimensional data manifolds including intrinsic dimension, curvature, density, and topological structure; the package is openly available at \url{https://pypi.org/project/perceptual-manifold-geometry/}.



\bibliographystyle{unsrtnat} 
\bibliography{ref}


\appendix

\end{document}